\documentclass[10pt,letterpaper]{article}
\usepackage[top=0.85in,left=1.0in,footskip=0.75in]{geometry}

\usepackage{amsmath,amssymb}

\usepackage{changepage}

\usepackage[utf8x]{inputenc}

\usepackage{textcomp,marvosym}

\usepackage{cite}

\usepackage{nameref}

\usepackage{amssymb}
\usepackage{amsmath}
\usepackage{braket}

\usepackage{microtype}
\DisableLigatures[f]{encoding = *, family = * }

\usepackage[table]{xcolor}

\usepackage{array}

\newcolumntype{+}{!{\vrule width 2pt}}

\newlength\savedwidth

\newcommand\thickhline{\noalign{\global\savedwidth\arrayrulewidth\global\arrayrulewidth 2pt}%
\hline
\noalign{\global\arrayrulewidth\savedwidth}}

\raggedright
\usepackage[aboveskip=1pt,labelfont=bf,labelsep=period,justification=raggedright,singlelinecheck=off]{caption}

\makeatletter
\renewcommand{\@biblabel}[1]{\quad#1.}
\makeatother

\date{}

\usepackage{lastpage,fancyhdr,graphicx}
\usepackage{epstopdf}
\fancyheadoffset[L]{2.25in}
\fancyfootoffset[L]{2.25in}
\begin{document}
\vspace*{0.2in}

\begin{flushleft}
{\Large
\textbf\newline{Semantically Enhanced Dynamic Bayesian Network for Detecting Sepsis Mortality Risk in ICU Patients with Infection} 
}
\newline
\\
Tony Wang, PhD\textsuperscript{1},
Tom Velez, PhD \textsuperscript{2},
Emilia Apostolova, PhD\textsuperscript{3*},
Tim Tschampel\textsuperscript{2},
Thuy L. Ngo, DO, MEd\textsuperscript{4},
Joy Hardison, MD MPH\textsuperscript{5}\\

\bigskip
\textbf{1} IMEDACS, Ann Arbor, MI, USA
\\
\textbf{2} Vivace Health Solutions Inc, Cardiff, CA, USA
\\
\textbf{3} Language.ai, Chicago, IL, USA
\\
\textbf{4} Johns Hopkins University, Baltimore, MD, USA
\\
\textbf{5} Scripps Memorial Hospital - Encinitas, Encinitas, CA, USA
\\
\bigskip

%
%
%
%
%
%

\end{flushleft}
\section*{Abstract}

Although timely sepsis diagnosis and prompt interventions in Intensive Care Unit (ICU) patients are associated with reduced mortality, early clinical recognition is frequently impeded by non-specific signs of infection and failure to detect signs of sepsis-induced organ dysfunction in a constellation of dynamically changing physiological data. 
The goal of this work is to identify patient at risk of life-threatening sepsis utilizing a data-centered and machine learning-driven approach. We derive a mortality risk predictive dynamic Bayesian network (DBN) guided by a customized sepsis knowledgebase and compare the predictive accuracy of the derived DBN with the Sepsis-related Organ Failure Assessment (SOFA) score, the Quick SOFA (qSOFA) score, the Simplified Acute Physiological Score (SAPS-II) and the Modified Early Warning Score (MEWS) tools. 

A customized sepsis ontology was used to derive the DBN node structure and semantically characterize temporal features derived from both structured physiological data and unstructured clinical notes. We assessed the performance in predicting mortality risk of the DBN predictive model and compared performance to other models using Receiver Operating Characteristic (ROC) curves, area under curve (AUROC), calibration curves, and risk distributions.

The derived dataset consists of 24,506 ICU stays from 19,623 patients with evidence of suspected infection, with 2,829 patients deceased at discharge. The DBN AUROC was found to be 0.91, which outperformed the SOFA (0.843), qSOFA (0.66), MEWS (0.729), and SAPS-II (0.766) scoring tools. Continuous Net Reclassification Index and Integrated Discrimination Improvement analysis supported the superiority DBN with respect to SOFA, qSOFA, MEWS, and SAPS-II.
Compared with conventional rule-based risk scoring tools, the sepsis knowledgebase-driven DBN algorithm offers improved performance for predicting mortality of infected patients in intensive care units.



\section*{Introduction}
In the U.S., up to 52\% of all hospital deaths, typically in the ICU, involve sepsis\cite{thiel2010early}.  Notably, patients presenting with initially less severe sepsis account for a majority of these sepsis deaths\cite{liu2014hospital}. 

Given this prevalence, predictive tools that identify at risk patients during early stages of disease progression could drive important reductions in hospital mortality. Studies indicate that patients often have detectable signatures of physiological decompensation or deterioration in monitoring data hours before events such as septic shock or unexpected death\cite{henry2015targeted,cretikos2007objective,churpek2014using,moss2016signatures}. 

At the same time, the clinical manifestations of sepsis are highly dynamic, depending on the initial site of infection, the causative organism, the pattern of acute organ dysfunction, the underlying health status of the patient, and the interval before initiation of treatment\cite{angus2013severe}. Physiological data tracking tools with excellent predictive value that assist in the timely identification of patients at imminent risk (with hours of an event) may lead to improved outcomes \cite{pmid16480490}.  Each hour of delay in the administration of recommended therapy is associated with a linear increase in the risk of mortality\cite{kumar2006duration,han2003early}.  

Tracking tools can also help the treatment team strike an effective balance between judicious utilization of limited, high cost monitoring resources while providing the highest intensity of care to optimize the patient’s survival\cite{buist2002effects,mao2012integrated}.

\subsection*{Sepsis Mortality Risk Prediction Tools}

The 2016 Third International Consensus Definitions for Sepsis Task Force defined sepsis as a “life-threatening organ dysfunction due to a dysregulated host response to infection”\cite{singer2016third}. Using a validation cohort of 7,932 ICU encounters with signs of infection and associated organ dysfunction and mortality observations (16\%), the Task Force retrospectively established the predictive validity of a change of two or more points in the SOFA score (AUROC = 0.74; 95\% CI, 0.73-0.76) for the identification of patients at risk of hospital mortality from sespsis using patient physiological data from up to 48 hours before to up to 24 hours after the onset of infection \cite{seymour2016assessment}. 

A qSOFA score, as a simple surrogate for organ dysfunction in cases lacking adequate SOFA physiological, data was similarly derived and validated (AUROC = 0.66; 95\% CI, 0.64-0.68), and compared with the Systemic Inflammatory Response Syndrome (SIRS) (AUROC = 0.64; 95\% CI, 0.62-0.66)\cite{seymour2016assessment}. Recent studies indicate that the use of two-point SIRS criteria, or two-point qSOFA criteria for identifying at risk ICU patients is not as effective as SOFA, thus limiting their use for predicting mortality in this setting\cite{freund2017prognostic,besen2016sepsis,raith2017prognostic}.

\subsection*{ICU Mortality Risk Prediction Tools}

A number of ICU patient scoring systems for assessing disease severity that predict mortality outcomes have been developed and used for standardizing research and for comparing the quality of patient care across ICUs. In addition to SOFA\cite{vincent1996sofa}, these tools include the SAPS\cite{le1993new}, Acute Physiologic and Chronic Health Evaluation (APACHE)\cite{knaus1991apache}, and the Mortality Prediction Model (MPM)\cite{lemeshow1993mortality}. 

No single instrument has convincing or proven superiority to another in its ability to predict death\cite{nassar2012caution,poole2012comparison,kuzniewicz2008variation}. For comparative purposes in this study, beyond SOFA, we have chosen to use SAPS-II, which, compared to APACHE, conveniently predicts mortality via a closed form equation. We also included the relatively simple (Modified Early Warning Score) MEWS\cite{gardner2006value}, initially proposed as a scoring method for emergency departments, that has been used for predicting length of stay and mortality in ICU setting\cite{reini2012prognostic}.

\subsection*{Machine Learning (ML)}

In addition to individualized patient rule-based analysis of clinical data, modern EMRs include the ability to analyze large repositories of institutional multi-patient longitudinal data with powerful ML algorithms that can learn complex patterns in physiological data without being explicitly programmed, unlike simpler rule-based systems. Most importantly, these algorithms can be trained to recognize/classify early patterns in large data repositories to identify with high precision  hospitalized patients at risk of sepsis mortality early enough to effectively intervene before clinical deterioration\cite{desautels2016prediction,mani2014medical,horng2017creating,taylor2016prediction}.

Generalized linear models (i.e. logistic regression, Cox regression), characterized as having high explanatory power, are the most common algorithms used to develop risk prediction models using EMR data\cite{goldstein2017opportunities}. Commonly used linear ML methods in sepsis predictive analytics include multivariate linear regression\cite{danner2017physiologically}, logistic regression\cite{schlapbach2017prediction,haskins2016predictors,raith2017prognostic,moss2016signatures} and linear support vector machine (SVM)\cite{horng2017creating,houthooft2015predictive} classifiers. However, sepsis has been characterized as a “complex systems” class problem\cite{mann2013complex}, with multiple organ failures that represent a “chaotic adaptation” to severe stress\cite{kilic2009sepsis} which cannot be explained by linear statistical systems\cite{saliba2008chaotic}. The recognized complexity of the disease where actual covariate interrelationships can be complex and non-linear may explain the modest predictive performance of tools derived using logistic regression. 

With the tremendous growth in the data science field, many nonlinear machine learning algorithms, although more “black box”, are now available. Not limited by linearity assumptions, they can explore different covariate interrelationship options for predicting mortality. Tang et al\cite{tang2010non} explored the use of a nonlinear support vector machine using features extracted from spectral analysis of cardiovascular monitors. Quinn et al\cite{quinn2009factorial} proposed a factorial switching linear dynamic system to model the patient states based on 8 physiological measurements. The 2012, PhysioNet/Computing in Cardiology Challenge aimed to tackle the problem of in-hospital mortality prediction using the MIMIC-II (Multiparameter Intelligent Monitoring in Intensive Care)\cite{saeed2011multiparameter} dataset with 36 physiologic time series from first 2 days of admission.  Challenge participants explored various prediction algorithms including time series motifs, neural networks, and Bayesian ensembles but show contradictory relative predictive performance results\cite{lee2012imputation,xia2012neural,johnson2012patient,mcmillan2012icu}. 

On the other hand, studies that have explored dynamic Bayesian networks that model temporal dynamics of features have shown promise\cite{nachimuthu2012early,sandri2014dynamic,peelen2010using}. Recently, a “super-learner ensemble” algorithm designed to find optimal combinations of a collection of prediction algorithms was reported to achieve enhanced performance in mortality prediction in ICU\cite{pirracchio2015mortality}. 

\subsection*{Dynamic Bayesian Networks (DBN)}

Another key characteristic of sepsis is the rapid progression of the disease\cite{bloos2014rapid}. Fast and appropriate therapy is the cornerstone of sepsis treatment\cite{srinivasan2012new}. Despite methodological advances in ML, most methods model temporal trends independently, without capturing correlated trends manifesting underlying changes in pathophysiologic states. A recent systematic review of risk prediction models derived from EMR data identified incorporation of time-varying (longitudinal) factors as a key area of improvement in future risk prediction modelling efforts\cite{goldstein2017opportunities}. Most machine learning algorithms take standalone snapshots of numerical measurements as features, mainly because they can be easily extracted and may have robust statistical properties. 

It is generally recognized that temporal trends, particularly in one or more observations reflecting a rapidly changing patient state, can be more expressive and informative than individual signs. Tools that provide bedside clinicians with information about a patient’s changing physiological status in a manner that is easy and fast to interpret may reduce the time needed to detect patients in need of life-saving interventions. 

To address these challenges, we investigate DBNs which have the ability to model complex dynamics of a system by placing correlated temporal manifestations into a network. DBNs have been used to model temporal dynamics of sepsis, to predict sepsis in presenting Emergency Department patients\cite{nachimuthu2012early}, and to predict sequences of organ failures in ICU patients\cite{sandri2014dynamic,peelen2010using}.

\subsection*{Concept Maps (Cmap)}

Our DBN approach applies an evidence-based network structure based on expert-constructed concept maps (Cmaps) which are a relatively recent semantic web technology\cite{novak1991twelve} that employs a user-friendly approach to enable generation of customized hierarchical ontologies in the form of concept graphs. 

Cmaps are semi-formal (subject concept-predicate-object concept triples) graphical expressions of diagnostic clinical reasoning as expressed by clinicians over observed evidence such as vials, labs and exam findings. Cmaps formalize “deep semantic” models reflecting clinical reasoning that can be used to contextually characterize raw patient encounter clinical history, time-tagged symptoms and signs, laboratory tests, medications, procedures, orders and other clinical data. The “structured reasoning” knowledge expressed in these graphs
includes both explicit evidence-based relationships (e.g. guidelines, protocols) between
concepts as well as implicit expertise (e.g. heuristics, atypical patterns) learned from
years of medical practice. We use these maps to create computable ontologies that can
be used to automatically semantically classify raw medical record data in effect
contextually interpreting real-time data based on this expert knowledge. In the case of
sepsis these models can be used to express concept/relationships across risk factors,
treatments, history and label otherwise ambiguous features such as vitals and labs which
are reflected in the DBN as causal nodes and relationships. 

In this study, customized high level concepts and relationships between concepts in the form of simple “subject-predicate-object” collections of sepsis domain nodes and relationships are enhanced with historically available medical ontologies such as the Systematized Nomenclature of Medicine (SNOMED)\footnote{\copyright The International Health Terminology Standards Development Organisation (IHTSDO)}. Cmaps are used to guide the DBN node structure and the contextual meaning of raw clinical data and to enable inductive rule-based decision support and clinical event characterizations for machine learning closely reflecting how expert clinicians would cognitively process patient data.

\begin{figure}[!h]
	\centering
	    \includegraphics[width=1.0\textwidth]{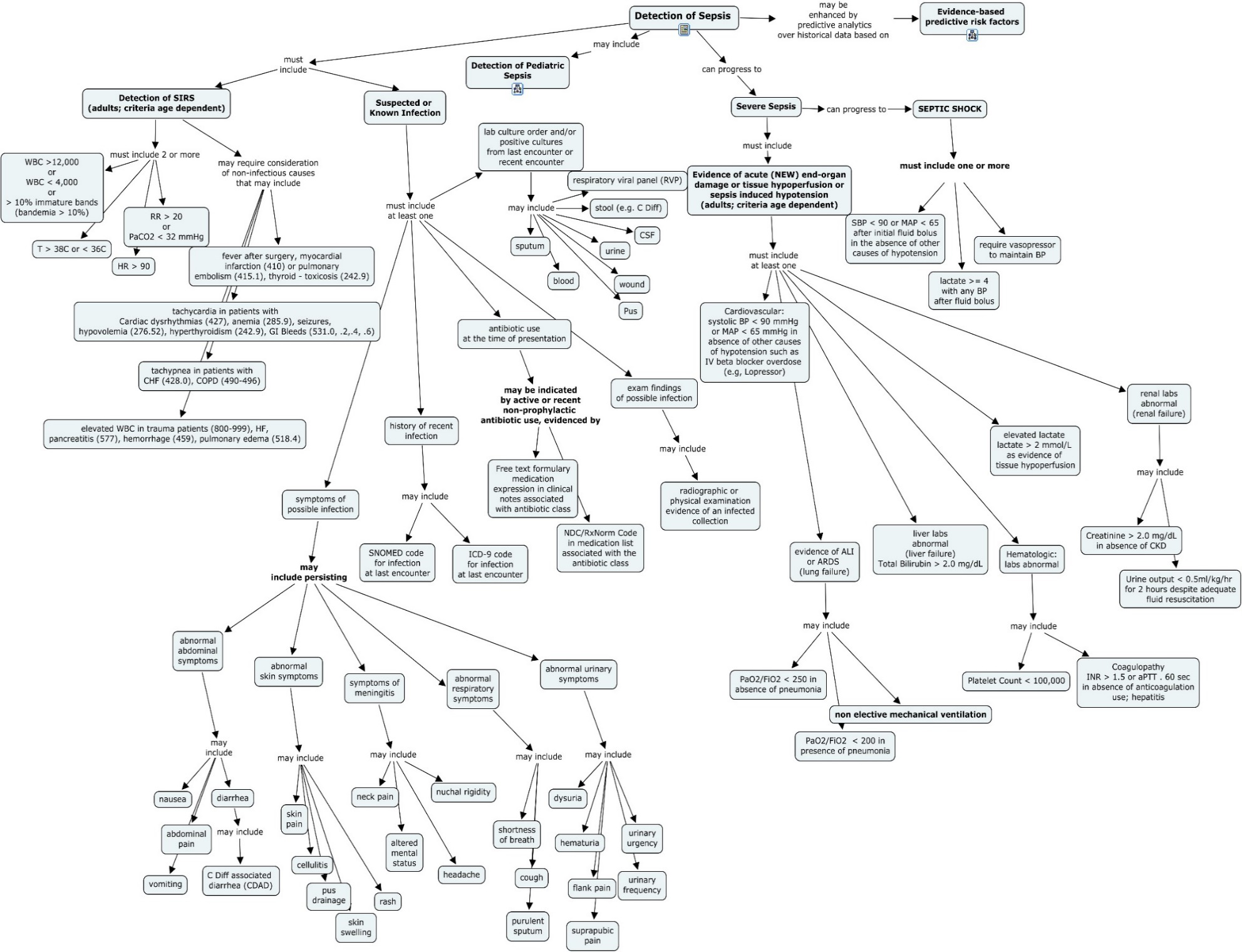}
\caption{{\bf Partial view of a sepsis concept map.}
Expert-constructed concept maps (Cmaps) representing evidence and clinician's reasoning for Sepsis detection (partial view).}
\label{fig1}
\end{figure}

Fig~\ref{fig1} shows a partial hierarchical Cmap representing high level sepsis (Sepsis-2) evidence\cite{dellinger2013surviving} and reasoning concepts. 

Although ontologies have been used in biomedical research\cite{hoehndorf2015role}, and there is growing interest in the use of semantic models in healthcare\cite{goossen2014detailed}, few existing applications of  predictive analytics in clinical decision support leverage the power of comprehensive customized ontologies in rule engine and machine learning applications. It has also been demonstrated that using Cmaps to elicit content knowledge from physicians can significantly reduce the time and costs necessary to implement a physician's knowledge/reasoning logic into operational systems\cite{brewer2012application}. The need for adaptation of clinical prediction models for application in local settings and across new problem domains is well recognized\cite{kappen2012adaptation}. The use of semantic knowledge graphs authored by expert clinicians in machine learning can be an effective mechanism for rapid validation of new models or recalibration of existing models in new settings and/or populations.

\section*{Materials and methods}
\subsection*{Study Dataset}

The data used in this study is exempted from IRB review, as Quorum Review IRB agrees that it meets the following criteria set forth in 
The Code of Federal Regulations, TITLE 45, Part 46.101(b)(4)\footnote{\url{https://www.hhs.gov/ohrp/regulations-and-policy/regulations/45-cfr-46/index.html}}:
 
\textit{Research involving the collection or study of existing data, documents, records,
pathological specimens, or diagnostic specimens, if these sources are publicly available
or if the information is recorded by the investigator in such a manner that subjects
cannot be identified, directly or through identifiers linked to the subjects.}

Clinical encounter data of adult patients (age\(>= \)18 years) were extracted from the MIMIC-II version 26 ICU database\cite{saeed2011multiparameter}. MIMIC-II consists of retrospective hospital admission data of patients admitted into Beth Israel Deaconess Medical Center from 2001 to 2008. Included ICUs are medical, surgical, trauma-surgical, coronary, cardiac surgery recovery, and medical/surgical care units.  Although MIMIC-II includes both time series data recorded in the EMR during encounters (e.g. vital signs/diagnostic laboratory results, free text nursing notes/radiology reports, medications, discharge summaries, treatments, etc.) as well as high-resolution physiological data (time series / waveforms) recorded by bedside monitors, only the time series data recorded in the EMR was used in this study.

\subsection*{Patient Inclusion}

Although MIMIC-II includes all patients who were admitted in the period, we focus on patients who experienced at least one ICU stay. Only patients who were determined to have had an infection or suspected infection were included in this analysis. Infection status was determined using a combination of structured fields and free-text nursing notes. We used the following criteria: (1) antibiotics and/or microbiological culture order during encounter; (2) ICD-9-CM diagnosis codes based on Sepanski et al. \cite{sepanski2014designing}; (3) free-text nursing notes containing text suggesting an infection or suspected infection.  

To identify patients with infection in nursing notes we developed a supervised ML algorithm utilizing a bag-of-words\cite{harris1954distributional} and linear kernel SVMs\cite{apostolova2017toward}.   The ML algorithm was trained on a dataset that was automatically generated using a simple heuristic. We observed that whenever there is an existing infection or a suspicion of infection, the nursing notes typically describe the fact that the patient is taking or is prescribed infection-treating antibiotics. Thus, identifying nursing notes describing the use of antibiotics will, in most cases, also identify nursing notes describing signs and symptoms of infection\footnote{While the MIMIC database contains structured medication information, the list of medications is associated with the patient's admission. An admission can contain numerous clinical notes and prescribed medications. This is why we were unable to use the structured medications data in order to construct a note-level training dataset}. Additionally, antibiotics were characterized as “prophylactic” (negating infection) based on reasoning over use patterns and patient context (e.g. pre- or post-surgery). We utilized word embeddings\cite{mikolov2013efficient} to create a list of rules that unambiguously identify the use of infection-treating antibiotics. Word embeddings were generated utilizing all available MIMIC-III nursing notes\footnote{We used vector size 200, window size 7, and continuous bag-of-words model.}. The initial set of antibiotics was then extended using the closest \textit{word embeddings} in terms of cosine distance.  For example, the closest words to the antibiotic \textit{amoxicillin} are \textit{amox, amoxacillin, amoxycillin, cefixime, suprax, amoxcillin, amoxicilin}. As shown, this includes misspellings, abbreviations, similar drugs and brand names. The extended list was then manually reviewed. The final infection-treating antibiotic list consists of 402 unambiguous expressions indicating antibiotics. 

Antibiotics, however, are sometimes negated and are often mentioned in the context of allergies (e.g. \textit{allergic to penicillin}). To distinguish between affirmed, negated, and speculated mentions of administered antibiotics, we also developed a set of rules in the form of keyword triggers. Similarly to the NegEx algorithm \cite{chapman2001simple}, we identified phrases that indicate uncertain or negated mentions of antibiotics that precede or follow a list of hand-crafted expression at the sentence and clause levels.

The described approach identified 186,158 nursing notes suggesting the unambiguous presence of infection (29\%) and 3,262 notes suggesting possible infection. The remaining 448,211 notes (70\%) were considered to comprise our negative dataset, i.e. not suggesting infection. The automatically generated dataset was used to train SVMs which achieved an F1-score ranging from 79 to 96\%\cite{apostolova2017toward}.

While we created a dataset relying on the mention of antibiotics, the ML algorithm was abled to identify positive infection-describing notes not containing mentions of antibiotics. Depending on the infection source and the specifics of the patient history, signs and symptoms of infection can vary widely. Utilizing ML enabled us to capture a wide range of infection signs and symptoms, often (but not always) accompanied by a description of antibiotics use. 

\subsection*{Data preparation}

DBN models a temporal process by discretizing time and building a network slice on each discretized time point on which the system’s status is presented. Basically, DBN network structure is repeated on the basis of time slices, and nodes on temporally different slices are connected by appropriate edges. Therefore, one important DBN data preparation step is to create aligned physiological data suitable for time slices across study patients with highly diverse vitals and lab measurement/observation rates. Fortunately, MIMIC-II represents an integrated set of diverse physiological measurements and events that are time-stamped over the entire ICU stay, enabling us to align measurements on a coherent time line and to locate an appropriate interval to which every measurement belongs.

Considering that the MIMIC-II sampling rate of physiologic vitals data varies from 12 to 120 minutes with a mean and median of approximately 60 minutes, a 1-hour interval between time-slices for vitals was found feasible. Lab measurements were far less frequent. To normalize variable measurement frequency, we used most recent readings for vitals and labs with a loopback window of 4 hours and 16 hours respectively, i.e., vitals such as systolic blood pressure (SBP) and temperature readings were considered “current” for 4 hours until a newer reading was recorded received. Similarly, lab results such as white blood cell (WBC) counts was considered “current” for 16 hours. To avoid effects of sparse vitals data, we tested 4-hour, 8-hour, and 12-hour intervals for mortality predictions.

A “rolled-back” approach was applied in selection of training data\cite{shavdia2007septic}, i.e. using the discharge time as a reference time of survival/mortality event, selecting time slices N hours prior to event for training. The rationale for applying this method is based on: (1) physiological measurements close to the ICU end event may have a stronger discriminating relationship with the event than more remote (earlier) measurements; (2) across the entire population of ICU patients, mortality is a statistically rare event. If time slices were to be selected clock-wise in the ICU encounter and limited to remote data, severe imbalance in death variable could bias the model to the majority class and have significant impact on mortality predictive model performance\cite{krawczyk2016learning}.

\begin{figure}[!h]
	\centering
		\includegraphics[width=1.0\textwidth]{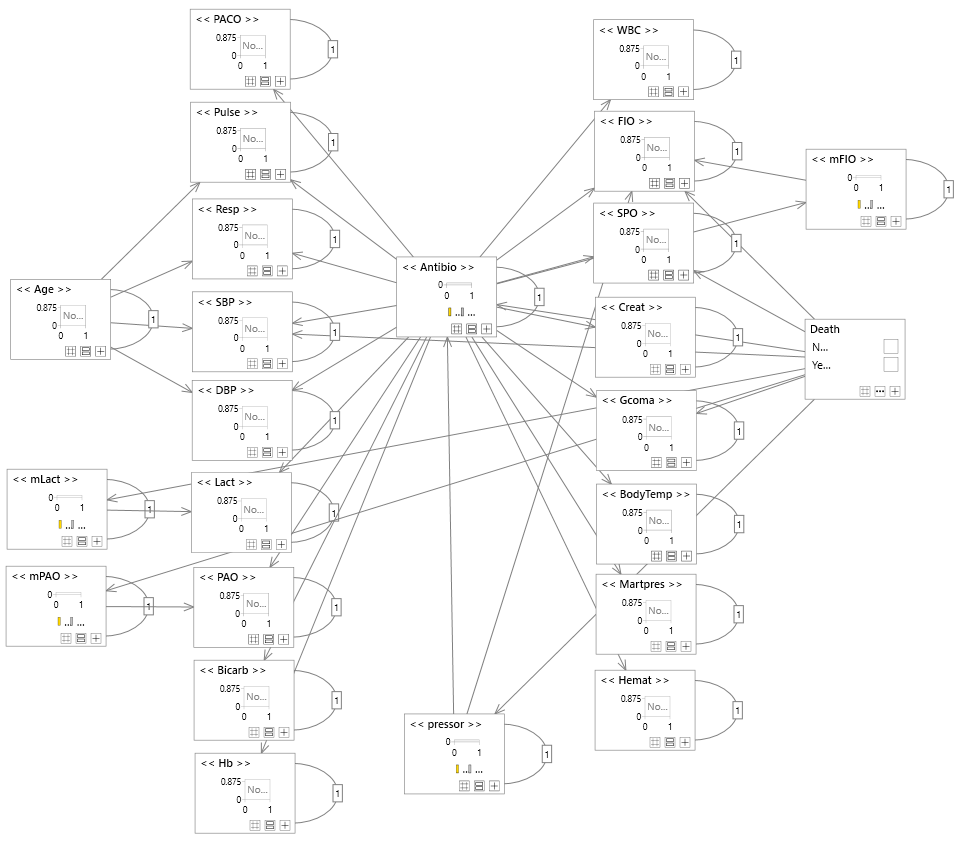}
\caption{{\bf Network Structure of DBN.}
Circle line with box 1 indicates the order 1 auto-regressive relationship between time slices. (Abbreviations used: ALT=Alanine Transaminase Test, AST=Aspartate Aminotransferase Test, DBP=Diastolic Blood Pressure, FiO2=Fraction of Inspired O2, GCS=Glasgow Coma Scale, INR=Prothrombin Time Test, MAP=Mean Arterial Pressure, PaCO2=Partial Pressure of Carbon Dioxide in Arterial Blood, PaO2=Partial Pressure of O2 in Arterial Blood, PlateletCnt=Platelet Count, SBP=Systolic Blood Pressure, SPO=Peripheral Capillary Oxygen Saturation, Uout=Urinary Output, WBC=White Blood Cell Count \textit{m} prior to a label indicates missing value, e.g. mFiO2=Missing Fraction of Inspired O2.)}
\label{fig2}
\end{figure}

Considered variables (shown as DBN nodes in Fig~\ref{fig2}) on a time slice are: 1) vital signs: heart rate, respiratory rate, body temperature, systolic blood pressure, diastolic blood pressure, mean arterial pressure, oxygen saturation, urinary output; 2) laboratory tests: white blood cell count, ALT/AST, bilirubin, platelets, hemoglobin, lactate, creatinine, bicarbonate; blood gas measurements (partial pressure of arterial oxygen, fraction of inspired oxygen, and partial pressure of arterial carbon dioxide); 4) Glasgow Coma Scale; and 5) indicators for non-prophylactic antibiotics and vasopressor usage.

Variables on a time slice are created by the following rules:
\begin{enumerate}
 \item Align all patient data by discharge timestamp, and look back 4 hours before discharge.
 \item For each vital sign, check the interval between discharge timestamp - 4 hours and discharge timestamp – 8 hours, to see if any measurement was taken. A variable will be missing if no measurement is taken during the interval. If a single measurement was taken, it’s treated as the value of a vital sign variable. If multiple measurements were taken during the 4-hour interval, a mean of these measurements is the value of a vital sign variable.
 \item For lab data, check the interval between discharge timestamp - 4 hours and discharge timestamp – 16 hours. Apply the same rules as those described above to find values for variables. That is, a physiological variable will be missing if no measurement is taken during the interval. If a single measurement was taken, it’s treated as the value of a physiological variable. If multiple measurements were taken during the 12-hour interval, a mean of these measurements is the value of a lab variable.
\end{enumerate}

To meet the normality assumption of continuous variables in a DBN, a normality transformation was applied to variables which are not normally distributed. Normalization was typically achieved using base-10 logarithmic transformation. In cases where log-transformation was not appropriate, the Box-Cox power transformation was applied\cite{box1964analysis}.  No special treatment for missing data was applied in our experiments. The Bayes Server\footnote{Version 7.8, Bayes Server Ltd. West Sussex, UK}, a commercially available suite of ML algorithms was used in our study to train and test the DBN models, supports parameter learning with missing values using the Expectation Maximization (EM) algorithm\cite{koller2009probabilistic}.

Vital and lab features were additionally semantically characterized using the ontology and associated rules to enhance specificity. For example, based on the finding from Lin and Haug\cite{lin2008exploiting}, missing indicators for lactate and/or blood gas parameters were added as features to model potential non-ignorable missingness. Additionally, abnormal lab values suggestive of sepsis-induced organ dysfunction were semantically viewed in the context of patient chronic conditions and/or use of medications that otherwise “explained” lab abnormalities and negated semantic assertions of infection etiology in the ontology.

\subsection*{Network Structure}

The basic DBN structure is directed by Cmaps with nodes in the DBN representing concepts in the Cmap, and edges between nodes representing connecting links on the Cmap. The process may be viewed as Cmap-bridged construction of an expert system\cite{almeida2014expert}. An unconstrained BN that is built from data starting with an empty network represents an NP-hard problem. The NP-hard problem is avoided with the use of Cmaps as the knowledge base\cite{dagum1993approximating,chickering2004large}.

A Bayesian Network (BN) is a graphical model describing the statistical relationships through complex interactions and dependencies among variables\cite{koller2009probabilistic}. The BN is composed of a Directed Acyclic Graph (DAG) structure \( G \) and distribution parameter \( \Theta\). A node \( X_i \) in \( G \) corresponds to a random variable, and an edge in \( G \) directed from node \( i \) to \( j \) encodes \( X_i \)’s conditional dependence on \( X_j \), with \( X_j \) being the parent of \( X_i \). By graphical model theory, a variable \( X_i \) is independent of its non-descendants given all its parents \( Pa(X_i) \) are in \( G \)\cite{koller2009probabilistic}. Then, the joint probability distribution over the variable set \( X=(X_1,...,X_m) \) can be decomposed by the chain rule:

\[ p(X)=\Pi_{i=1}^m p(X_i | Pa(X_i))  \]

The parameter set \( \Theta =\{\Theta_i\}_{i=1,..,n} \) specifies the parameters of each conditional distribution \( p(X_i|Pa(X_i)) \). The meaning of \( \Theta_i \)  is interpreted by specific distributions: \( \Theta_i \)  is simply a conditional probability table if the assumed distribution is multinomial, and \( \Theta_i \) is the mean and variance when the distribution is Gaussian.

The temporal dynamics of vital signs and lab measurements is modeled using the DBN framework. A DBN is an extended BN used to model temporal processes\cite{murphy2002dynamic}. In a DBN, random processes are represented by a set of nodes \( X_i[t] \), the random variable of process \( X_i \) at time \( t \). The edges within slice $t$ indicate the relationship between these random variables, and additional edges between $t$ and $t+1$ model the time dependencies. Usually, process \( X \) is assumed to be Markovian. Any node in a DBN is only allowed to be linked from the nodes in the same or previous slice, i.e. \( Pa(X_i[t] \in \{X[t-1],X[t]\}) \). DBN generalizes hidden Markov model (HMM) and state space model (SSM), by representing the hidden / observed state in terms of variables\cite{murphy2002dynamic}. The basic BN structure for body dynamics is extended hierarchically, such that the model allows for the key parameters of the model to differ among time slices to account for potential non-linearity over time, and across individuals to account for between-patient heterogeneity.

Fig~\ref{fig2} shows the DBN network structure of a time slice of physiological variables used for mortality prediction in this study. DBNs have the desirable property that they allow for interpretation of interactions between the different variables, and can be used to predict event in far future \cite{murphy2002dynamic}. As an example of
feature interactions we explored the relationship between acute respiratory distress and acute
kidney injury via links between creatinine and P/F factors in the model. With forward-backward operators, DBN can make online prediction of events on any future timeslice, which is similar to time-series models. In this study, we will utilize this property to examine its performance in temporally remote time frames.

\subsection*{Calibration}

Producing well-calibrated prediction probabilities is crucial for supporting clinical decision making. To make sure the predicted and recorded probabilities of death coincide well, we applied the method of ensembling linear trend estimation\cite{naeini2016binary} to calibrate the predicted probabilities of mortality. The method utilizes the \( \l_1 \)  trend filtering signal approximation approach to find the mapping from uncalibrated classification scores to the calibrated probability estimates.

\subsection*{Performance Measures}

We examine the capability of DBN in predicting ICU mortality by evaluating sensitivity, specificity, positive predictive value, negative predictive value, F1 measures, and AUROC obtained through 10-fold cross validation, reported with 95\% confidence interval obtained through a computationally efficient influence curve based approach\cite{ledell2015computationally}.  DBN’s performance is compared with SOFA, qSOFA, MEWS, and SAPS-II. For SOFA, we obtained mortality prediction by regressing ICU death indicator on SOFA scores (main effect only). Both the first SOFA score and maximal SOFA score are considered in separate models. The recommended cut-point 2 is used for calculating evaluation metrics for qSOFA\cite{raith2017prognostic}.  

To examine its performance in predicting events in relatively remote future, we considered two validation sets. In the first validation set, time slices are created clockwisely, that is, the first time slice started from the admission timestamp. with 3-timeslice data, we will predict in-ICU mortality (1) within 12 hours after the third timeslice, and (2) within 24 hours after the third timeslice. We name this dataset as `Validate Set 1`. The second validating set is created with rolled back-method. However, we aim to predicting mortality after discharge from ICU, instead of in-ICU mortality. Again, two predictions will be generated, within-12-hour mortality and within-24-hour mortality. All validation sets are composed of died patients in designed time frames and randomly selected 10\% survivors. For each prediction on the validation set, DBN is trained on the set (constructed with the rolled-back method described above) with patients in the validation set excluded based on patient ID and ICU-stay ID.

MEWS is constructed with a modification to the neurological component. The original component scores 0 for “alert”, 1 for “Reacting to voice”, 2 for “Reacting to pain”, and 3 for “Unresponsive”. In our construction, GCS is used with arbitarily chosen cut-offs which mimicks the relationship between GCS and head injury \cite{pmid17212177, pmid25111571}: neurological score =0 if 14≤GCS≤15, neurological score =1 for 10≤GCS≤14, neurological score =2 for 6≤GCS≤10, and neurological score =3 if GCS\(<\)6. Similar to SOFA, the mortality prediction is obtained by regressing ICU death indicator on MEWS (main effect only). 

For Simplified Acute Physiology Score II (SAPS-II), the predicted probability is obtained by formula which is suggested by Le Gall et al.\cite{le1993new}:

\begin{equation}
\log \left(\frac{P_{r(death)}}{1 - P_{r(death)}}\right) = -7.7631 + 0.0737 \times SAPSII + 0.9971 \times \log(1+SAPSII)
\end{equation}

ROC curves and boxplots of predicted probabilities of death for survivors and non-survivors are displayed for assisting the assessment of discrimination capability (Fig~\ref{fig3} and Fig~\ref{fig4}).

Summary reclassification measures of the continuous Net Reclassification Index (cNRI) and the Integrated Discrimination Improvement (IDI) will be calculated with SOFA or SAPS-II scoring classifier as the initial model and DBN as the updated classifier. Positive values of the cNRI and IDI suggest that the updated classifier has better discriminative ability than the initial classifier, whereas negative values suggest the opposite\cite{cook2007use,cook2008statistical,pencina2008evaluating}.

\section*{Results}

\subsection*{Study Data}

24,506 ICU stays for 19,623 adult patients with suspected or confirmed infection are included in the analysis out of a total of 29,431 ICU stays for 23,701 patients. Demographics and baseline characteristics of patients are shown in Table~\ref{tab1} indicating that 2,829 (14.4\%) patients were deceased at discharge. Non-survivors were older with median age 73.7 years (IQR: 59.5 – 82.9) than the survivors at 64.4 years median age (IQR: 50.7 – 76.9). Non-survivors stayed in the ICU a median 3.9 days (IQR: 1.8 – 8.5), compared to 2.3 median LOS (length of stay) for survivors (IQR: 1.3 - 4.7). Table~\ref{tab2} shows the descriptive statistics of vital signs and lab measurements. All statistical analyses are done with openly statistical computing software R and associated risk prediction modules (R Version 3.3.1, PredictABEL\cite{kundu2011predictabel}, pROC\cite{robin2011proc}, and caret\cite{kuhn2008caret}.)

\begin{table}[!ht]
\begin{adjustwidth}{0.25in}{0in} 
\centering
\caption{
{\bf Demographic and ICU encounter characteristics by mortality at discharge.}}
\begin{tabular}{l|l|l|l}
\hline \thickhline
Characteristics, N (\%) & Survivor & Non-survivor & Total \\
 & 16794 (85.6\%) & 2829 (14.4\%) & 19623 \\ \thickhline
\hline
Male, n (\%) & 9516(56.7) & 1519(53.7) & 11035(56.3)\\
\hline
\:\:\:\:\:\:\:\:Race, n (\%) & & &\\
White & 11910(70.9) & 1960(69.3) &13870(70.7)\\
Black & 1282(7.6) & 165(5.8) & 1447(7.4)\\
Hispanic & 536(3.2) & 45(1.6)& 581(3.0)\\
Asian & 366(2.2) & 57(2.0) & 423(2.2)\\
    Other & 2700(16.1) & 602(21.3) & 3302(16.8)\\
	\hline
\:\:\:\:\:\:\:\:Admission Type, n (\%) & & &\\
Elective & 3262(16.5) & 112(4) & 3374(14.9)\\
    Emergent & 15818(80) & 2587(91.6) & 18405(81.5)\\
    Urgent & 689(3.5) & 126(4.5) & 815(3.6)\\
	\hline
\:\:\:\:\:\:\:\:ICU Type, n (\%) & & & \\
    Coronary Care Unit & 3910(19.8) & 620(21.9) & 4530(20)\\
    Cardiac surgery recovery & 7001(35.4) & 638(22.6) & 7639(33.8)\\
    Combined Medical/Surgical ICU & 2150(10.9) & 320(11.3) & 2470(10.9)\\
    Medical ICU & 5640(28.5) & 1137(40.2) & 6777(30.0)\\
    Surgical ICU & 1068(5.4) & 110(3.9) & 1178(5.2)\\
 \hline
 Pressor usage, n (\%) &	4533(21.3) &	1573(49.0)&	6106(24.9)\\
 \hline
 Antibiotics usage, n (\%)&	10780(52.4)&	1863(58.0)&	13305(56.0)\\
  \hline
Age, median (IQR) & 64.4(50.7, 76.9) & 73.7(59.5, 82.9) & 65.8(51.9, 78.0)\\
\hline
LOS, median (IQR) & 2.3(1.3, 4.7) & 3.9(1.8, 8.5) & 2.5(1.3, 5.0)\\
\hline \thickhline

\end{tabular}
\begin{flushleft} IQR = Inter-quartile range, LOS = Length of Stay.
\end{flushleft}
\label{tab1}
\end{adjustwidth}
\end{table}

Tables \ref{tab2} and \ref{tab3} below shows measured physiological variables and vital signs at the time slice close to discharge, and comparative performances of ICU mortality predictions using SOFA, qSOFA, SAPS-II, MEWS, and the DBN.

\begin{table}[!ht]
\begin{adjustwidth}{-0.25in}{0in} 
\centering
\caption{
{\bf  Physiological measurements at the time slice closest to discharge: median (IQR).}}
\begin{tabular}{l|l|l|l}
\hline \thickhline
Variable & Survivor (Q1, Q3) & Non-survivor (Q1, Q3) & Total (Q1, Q3)\\ \thickhline
Diastolic Blood Pressure & 59.3(51.5, 68.3) & 47.7(36.7, 58.3) & 58.3(50.2, 67.5)\\
Systolic Blood Pressure & 122.25(110.0, 136.8) & 96.5(74.6, 117.7) & 120.5(107.3, 135.6)\\
Temperature & 36.8(36.4, 37.2) & 36.8(36.1, 37.6) & 36.8(36.4, 37.2)\\
PaCO2 & 41(37, 46) & 39(32, 48) & 41(36, 46)\\
Pulse & 82.3(72.6, 92.5) & 86.8(71.8, 104.0) & 82.7(72.5, 93.5)\\
Respiratory Rate & 19.27(16.8, 22.3) & 19.5(14.4, 24.3) & 19.3(16.5, 22.5)\\
Creatinine & 0.9(0.7, 1.2) & 1.6(0.9, 2.8) & 0.9(0.7, 1.3)\\
WBC & 10.1(7.6, 13.1) & 13.9(8.7, 20.3) & 10.2(7.7, 13.5)\\
FiO2 & 0.41(0.4, 0.5) & 0.5(0.4, 0.8) & 0.5(0.4, 0.6)\\
PaO2 & 100.0(82.0, 126.0) & 100.3(76.0, 146.0) & 100.0(80.0, 130.4)\\
SpO2 & 97(95.6, 98.3) & 93.8(84.8, 97.4) & 97(95.3, 98.3)\\
Lactate & 1.3(1, 1.8) & 5.65(2.72, 11.3) & 1.9(1.2, 5.3)\\
GCS & 15(15, 15) & 6.5(3, 11) & 15(14, 15)\\
Hemoglobin & 10.4(9.4, 11.5) & 10(9.0, 11.2) & 10.3(9.4, 11.4)\\
Bicarbonate & 26(23, 28) & 21(16, 26) & 25(23, 28)\\
Mean Arterial Pressure & 81.25(72.3, 92.5) & 63(50.0, 76.1) & 78.5(68.8, 90.5)\\

INR &	10.4(9.4, 11.45)&	10.1(9.01, 11.2)&	10.3(9.4, 11.4) \\
Platelet count &	1.3(1, 1.8)&	5.7(2.5, 11.8)&	2.07(1.3, 6.2) \\
ALT&	2.9(2.5, 3.3)&	2.5(2.1, 3)&	2.8(2.4, 3.3) \\
AST&	42(21, 98)&	82(25, 285)&	46(21, 127) \\
Bilirubin&	26(23, 28)&	22(17, 26.5)&	25(23, 28) \\
\hline \thickhline
\end{tabular}
\begin{flushleft} IQR = Inter-quartile range
\end{flushleft}
\label{tab2}
\end{adjustwidth}
\end{table}

\begin{table}[!ht]
\begin{adjustwidth}{-0.25in}{0in} 
\centering
\caption{
{\bf Comparison of performance of scores in mortality prediction among patients with infection. }}
\scriptsize
\begin{tabular}{l|llllll}
	
\hline
& qSOFA & SOFA First & SOFA Max & SAPSII & MEWS* & DBN \\ 
\hline
AUC & 0.66(0.637-0.683) & 0.756(0.73-0.781) & 0.853(0.816-0.871) & 0.766(0.735-0.798) & 0.729(0.686-0.771) & 0.913(0.906-0.919) \\
Sensitivity & 0.846(0.239-1.0) & 0.571(0.317-0.825) & 0.665(0.544-0.787) & 0.759(0.647-0.871) & 0.518(0.339-0.696) & 0.825(0.802-0.849) \\
Specificity & 0.288(0-0.874) & 0.767(0.51-1.023) & 0.818(0.707-0.929) & 0.63(0.536-0.724) & 0.775(0.608-0.943) & 0.874(0.84-0.908) \\
PPV & 0.104(0.076-0.133) & 0.248(0.085-0.412) & 0.301(0.21-0.392) & 0.163(0.131-0.195) & 0.19(0.104-0.277) & 0.474(0.416-0.533) \\
NPV & 0.975(0.931-1.018) & 0.942(0.923-0.961) & 0.956(0.942-0.97) & 0.966(0.955-0.977) & 0.945(0.934-0.955) & 0.973(0.971-0.976) \\
F1 & 0.187(0.146-0.228) & 0.326(0.224-0.428) & 0.41(0.341-0.478) & 0.268(0.224-0.312) & 0.272(0.199-0.344) & 0.602(0.56-0.643) \\
\hline 
\end{tabular}
\begin{flushleft} *Scoring of neurological variable in MEWS is based on GCS: neurological score =0 if 14≤GCS≤15, neurological score =1 for 10≤GCS≤14, neurological score =2 for 6≤GCS≤10, and neurological score =3 if GCS \(<\) 6.
\end{flushleft}
\label{tab3}
\end{adjustwidth}
\end{table}

\begin{figure}[!h]
	\centering
		\includegraphics[width=0.75\textwidth]{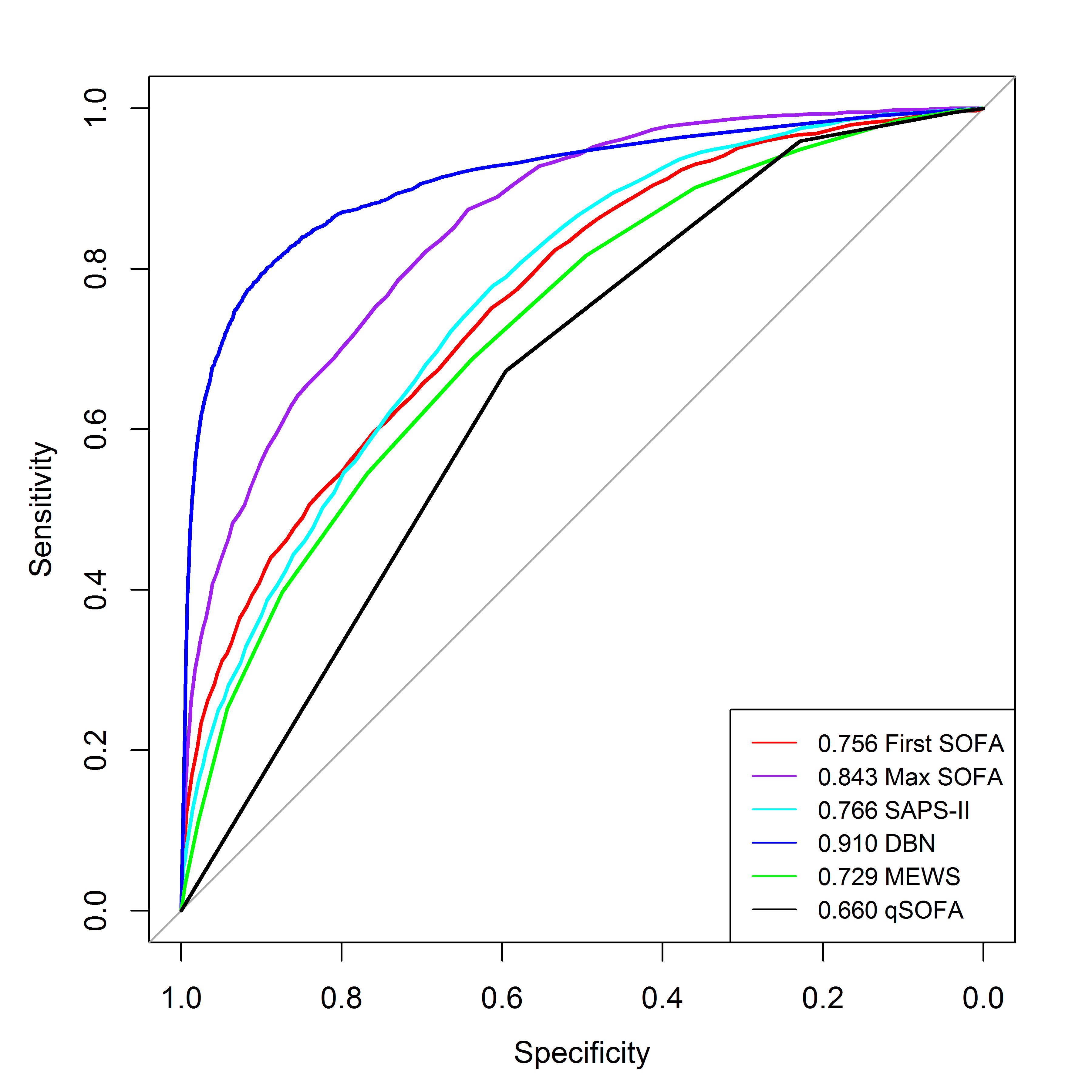}
\caption{{\bf Receiver-operating characteristic curves of all scoring methods.}}
\label{fig3}
\end{figure}

Fig~\ref{fig3} plots ROC curves of all scoring methods. The AUROC is 0.66 (95\% CI: 0.637 – 0.683) for qSOFA, 0.756(95\% CI: 0.73 - 0.781) for the first SOFA score, 0.843 (95\% CI: 0.816 - 0.871) for the maximum SOFA score, 0.729 (95\% CI: 0.686 – 0.771) for MEWS, and 0.766 (95\% CI: 0.735 - 0.798) for the SAPS-II score. A substantial improvement in performance is observed from the newly built DBN model, which gives us an AUROC of 0.91(95\% CI: 0.888 - 0.933). Table \ref{onlinepred} indicates great predicting capabilities for both in-ICU and after-discharge mortality, especially for the time relatively close to the last time point set for training. As expected, reductions in performance are observed when the time is relatively far away, e.g. for mortality within 24 hour after discharge, AUC decreased from 0.968 (95\% CI: 0.958 - 0.978) to 0.866 (95\% CI: 0.839 - 0.893). Cox calibration analysis suggests that the DBN prediction is well calibrated (\( \alpha = 0.011, \beta = 1.008, U = -7.57 \times 10^-5 , p<0.8676 \), see \nameref{S1_Appendix} for more details).

\begin{table}[!ht]
	\begin{adjustwidth}{0in}{0in} 
		\centering
		\caption{{\bf Performance of Online inference.}} \label{onlinepred}
		\begin{tabular}{l|l|ll}	
			\hline
			Time	& Metric   & In-ICU Mortality & After-discharge Mortality   \\
			\hline
			12 hours & AUC & 0.985(0.981 - 0.989)  & 0.968(0.958 - 0.978)  \\
					& Sensitivity & 0.940(0.899 - 0.969)  &  1(1 - 1) \\
					& Specificity & 0.942(0.923 - 0.977 & 0.941(0.937 - 0.957)  \\
					& & & \\
			24 hours & AUC & 0.975(0.967 - 0.982) & 0.866(0.839 - 0.893) \\
					 & Sensitivity & 0.923(0.882 - 0.952) & 0.721(0.655 - 0.819) \\
					 & Specificity & 0.940(0.918 - 0.977) &  0.889(0.783 - 0.912) \\
			\hline 
		\end{tabular}
		\begin{flushleft}
		\end{flushleft}
		\label{tab4}
	\end{adjustwidth}
\end{table}

\begin{figure}[!h]
	\centering
		\includegraphics[width=0.9\textwidth]{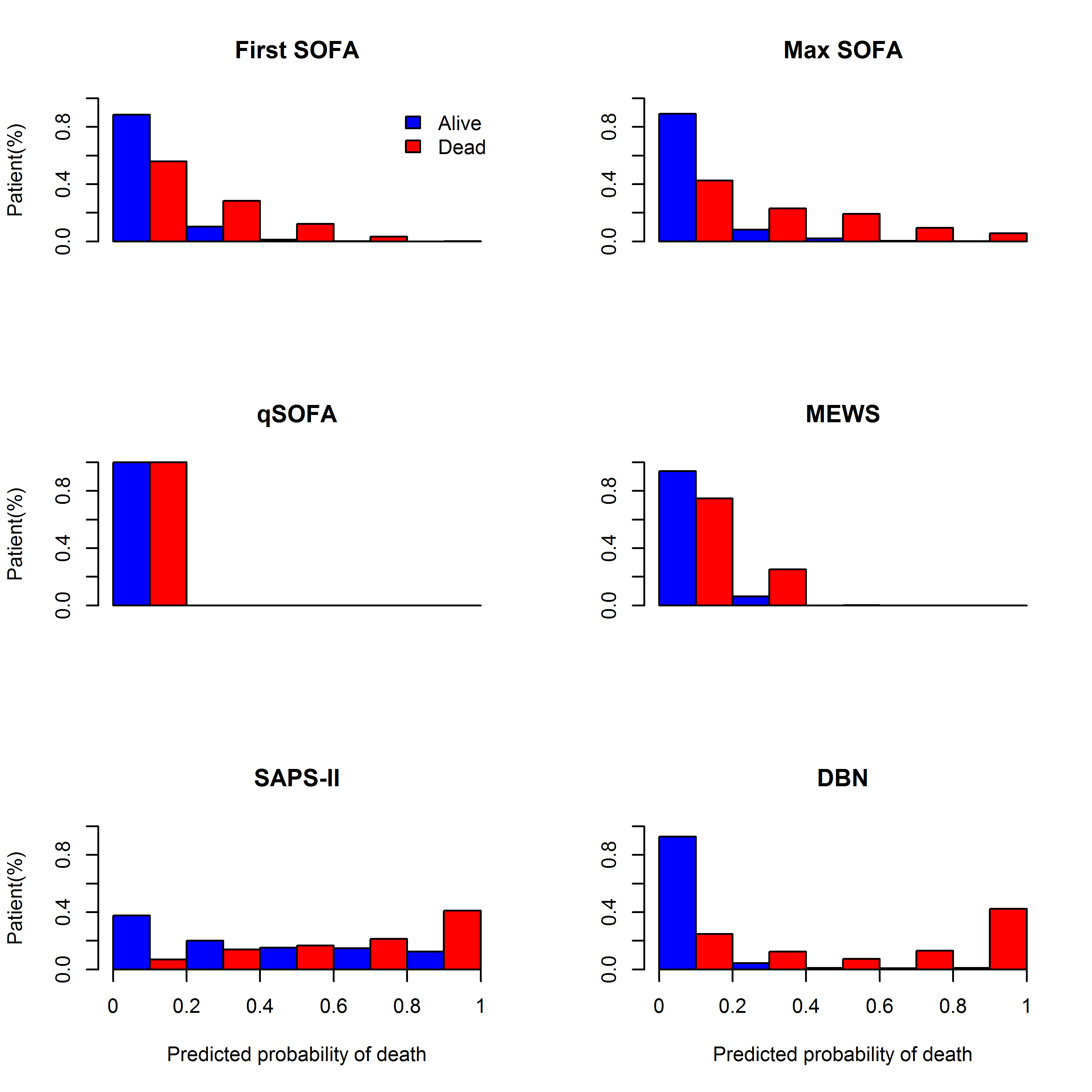}
\caption{{\bf Distribution of the predicted probability of death in the non-survivors and survivors.}}
\label{fig4}
\end{figure}

We tested whether parameter estimates learned with the addition of extra slices of data could improve the prediction performance of the DBN and determined that injecting three slices of data is sufficient to achieve the best performance (Table~\ref{tab5}). No significant change can be seen when adding another slice. Table~\ref{tab6} clearly indicates that DBN has significant improvement over SOFA, qSOFA, MEWS, or SAPS-II scores, since all means of NRI and IDI are positive and no 95\% confidence intervals include zero. Distributions of predicted probabilities of death are plotted in Fig~\ref{fig4} with respect to survivorship status. For the first SOFA, MEWS, and SAPS-II, great proportions of false positives and false negatives are observed regardless of the choice of cut-off point. There are similarities between max SOFA and DBN in the distributions of the predicted probabilities of death, while DBN has reduced proportion of false positive predictions.

\begin{table}[!ht]
\begin{adjustwidth}{0in}{0in} 
\centering
\caption{
{\bf Effect of adding slices: network with 4-hour rolled-back. }}
\begin{tabular}{l|lll}	
\hline
 &  2 slices & 3 slices & 4 slices \\
 \hline
  AUC & 0.856(0.826 - 0.885) & 0.91(0.888 - 0.933) & 0.901(0.888 - 0.915) \\
   Sensitivity & 0.749(0.668 - 0.829) & 0.814(0.773 - 0.856) & 0.814(0.746 - 0.881) \\
    Specificity & 0.839(0.801 - 0.877) & 0.880(0.851 - 0.909) & 0.886(0.843 - 0.928) \\
     PPV & 0.376(0.340 - 0.412) & 0.492(0.436 - 0.549) & 0.483(0.407 - 0.559) \\
     NPV & 0.963(0.953 - 0.973) & 0.971(0.965 - 0.977) & 0.974(0.967 - 0.981) \\
      F measure & 0.500(0.470 - 0.530) & 0.613(0.571 - 0.655) & 0.605(0.558 - 0.651) \\
\hline 
\end{tabular}
\begin{flushleft}
\end{flushleft}
\label{tab5}
\end{adjustwidth}
\end{table}

\begin{table}[!ht]
\begin{adjustwidth}{0.0in}{0in} 
\centering
\caption{
{\bf Reclassification statistics with respect to DBN as the updated classifier: mean (95\% CI). }}

\begin{tabular}{l|ll}
	
\hline
  
 Initial classifier & Continuous NRI & IDI \\
  \hline
  First SOFA & 1.23 (1.20 - 1.34) & 0.48 (0.47 - 0.50) \\
  Maximal SOFA & 1.20 (1.17 - 1.24) & 0.41 (0.39 - 0.42) \\
  SAPS-II & 1.09 (1.05 – 1.12) & 0.34 (0.33 - 0.36) \\
  MEWS & 1.23 (1.20 – 1.26) & 0.52 (0.51 – 0.54) \\
   qSOFA & 1.24 (1.21 – 1.28) & 0.55 (0.54 – 0.57)\\
  
\hline 
\end{tabular}
\begin{flushleft}NRI = Net Reclassification Index; IDI = Integrated Discrimination Improvement
\end{flushleft}
\label{tab6}
\end{adjustwidth}
\end{table}

\section*{Discussion}

We developed a DBN based algorithm for scoring the probability of death of ICU patients using `near event` data. The endpoint of in-hospital mortality has become the standard metric for early warning systems assessing risk for yet-to-be identified sepsis patients in the ICU. ICU  mortality risk prediction tools such as SOFA, SAP-II, and MEWS have been available for some time and have been of limited assistance to the management of individual patients. With the new Sepsis-3 standards published by The Sepsis Definitions Task Force\cite{singer2016third} for identifying at risk patients for this potentially deadly disease, SOFA has taken on a new importance in the ICU as a basis for decision support tools designed to expedite sepsis treatments and improve outcomes. The new definition calls for the detection of a simple change in SOFA score of ≥ 2 points from baseline to identify at risk patients suspected of having an infection and is the basis of many alerting tools. Knowing the highly nonlinear, chaotic nature of sepsis, we believe that a more specific, sophisticated method employing machine learning using time series data can more accurately and precisely assess individualized patient risk. Our DBN model achieves this and, by accurate predictions hours before events, may potentially contribute to efforts by bedside and discharge clinicians to reduce sepsis mortality.

Three aspects may explain the improved performance of the DBN model. 

First, compared to static parametric models commonly used for machine learning in medicine (e.g. logistic regression), the nonlinear temporally dynamic relationships between variables in a complex system such as the human body responding to an infection in a dysregulated manner\cite{singer2016third} may be more adequately captured by a highly interconnected DBN network structures that explicitly model the temporal dynamics of features. 

Second, beyond the overall structure of the DBN, the ontological models used in this study provided semantic characterizations of raw EMR data that were then used as training features. For example, the ontology characterized recorded events such as administration of a specific medication (e.g. cefazolin) in a patient with a related event “surgery” and no other evidence of infection as a “prophylactic antibiotic” distinguished infected from non-infected patients. Similarly, semantic characterizations of events such as abnormal vitals immediately following surgery or abnormal labs otherwise “explained by” medications (e.g. elevated INR following anticoagulant use) or chronic conditions distinguished septic-induced acute organ dysfunctions from chronic conditions.  We believe these reductions of “ambiguous feature noise” served to improve recognition of valid physiological data signatures associated with mortality outcomes. 

Third, unlike variable transformation (giving scores based on different thresholds) used by SOFA, MEWS, or SAPS scoring approaches, continuous versions of variables are used in DBN parameter learning. The threshold-based data transformation method may lead to information loss, indicated by the results from an experiment completed by Pirracchio et al\cite{pirracchio2015mortality}. As a side experiment, they compared the performance of original SAPS-II with a logistic model using all untransformed SAPS-II predictors as independent variables, and found the prediction from the logistic model to be better.

Another advantage of the study is that all ICU stays are included in this analysis. It is possible that measurements from multiple ICU stays of a single patient may be strongly correlated. Such correlation may have potential impact on model performance if it is not taken into consideration\cite{verbeke2014analysis}. This is one limitation identified by the Super ICU Learner Algorithm (SICULA) project\cite{pirracchio2015mortality}. However, the performance of DBN suggests the impact is limited. Compared to the SICULA study in which only single stay from each patient was included, this DBN with all ICU stays still outperforms SICULA.

It should be noted that models such as SOFA were not designed to be implemented by machines/EHRs, but rather these are bedside tools designed to be used by the clinician. When making evaluations, clinicians consider all additional factors of each individual patient. For example, the INR being elevated because of warfarin use, not because of sepsis and DIC. Such factors are not typically accounted for by the existing scoring tools. 

SOFA criteria for organ dysfunction associated with sepsis is defined as an “acute change in total SOFA score >= 2 points consequent to infection”, which requires evaluation of 2 key concepts which cannot be directly clinically measured: “acute change” and “consequent to”.  Clinicians can use complex reasoning and experience over contextual knowledge of a patient clinical history, risk factors and current treatments to adjudicate if the presence of SOFA score >= 2 points in a time series of physiological data is in fact an “acute change” from some patient “baseline” and this this change is “consequent to” an infection leading to a diagnosis of sepsis as opposed to a myriad of other possible non-infectious explanations for the change. This reasoning may be based on a blend of evidence-based rules as well as years of clinical experience with similar change patterns in infected patients.  DBN models can capture the temporal dynamics over large populations enhancing rule-based tools with experiential knowledge derived by learning over large datasets.

The clinical utility of a high precision `near event` predictive model is unknown and will require evaluation in pilot studies. As suggested, a tool such as the derived DBN may be useful to discharge clinicians in identifying patients near ICU discharge that are more prone to readmission or death and inform enhanced transfer decisions. Another limitation of the study is that we used SOFA and SAPS-II as reference scores while updated versions are available. This is partly due to a limitation of MIMIC-II. Some of the predictors included in the updated version of scores, e.g. SAPS-III and APACHE-III, are not directly available in the MIMIC-II database. However, it has been noted that the updated versions of scores are associated with similar drawbacks as the old version\cite{poole2012comparison,sakr2008comparison}. 

Another drawback of MIMIC-II is that all samples come from a single hospital, which may lead to limited generalizability due to limited case heterogeneity. However, our Cmap-driven DBN concept embraces iterative intra-institutional continuous improvement (retraining) of the model as new evidence, experience and new data dictates, and calibration/re-validation in new institutional settings. Cmaps can be revised and refined by clinicians using web-based tools to iteratively improve semantic model performance in interpreting patient data, with consequential benefits to subsequent ML. Additionally, as noted above, data driven-based predictive models work best in derived populations. The reuse of ontological structures combined with availability of powerful arrays of cloud-based cluster computing resources may soon enable the concept of “near real-time machine learning for re-calibration/validation” using new patient encounter datasets as they occur in real time.

In summary, we constructed a DBN based on clinician constructed Cmaps and found that its ability to predict mortality outperforms the more conventional scores SOFA, qSOFA, MEWS, SAPS-II and the more recently proposed SICULA model. In future work, we plan to confirm the utility of our highly predictive `near event` model as a
decision support tool and validate the algorithm using external data for improved case
heterogeneity. While additional validation work remains to be done, the promising DBN algorithm could be valuable as an alternative or addition to conventional scores in clinical setting.

\bibliography{plos}
\bibliographystyle{plos2015}

\newpage
\section*{Appendix 1. Calibration plots}

Cox calibration method is applied to assess calibration property of each score. Denote by  \( \hat{p_i} \) the predicted probability of mortality from each scoring algorithm, then Cox calibration regress the observed log-odds of mortality on the predicted log-odds:

\[ log \{ \dfrac{P_r (Y=1)}{P_r (Y=0)} \}= \alpha + \beta log \{ \dfrac{\hat{p_i}}{1-\hat{p_i}} \}   \]

Where \( \alpha \)  and \( \beta \)  are coefficients. When there is a perfect calibration between the observed probability and predicted probability, \( \alpha=1 \)  and \( \beta=1 \) .  Thus, the estimated \( \hat{\alpha} \)  and \( \hat{\beta} \)  based on data will be tested with respect to the null hypothesis  \( H_0: ( \alpha  \beta)=(0  1) \) using a U-statistic, to see the degree of deviation from ideal calibration.

Among all scoring methods, only SAPS-II shows severe deviation from ideal calibration. Its estimated \( \hat{\alpha}=-2.314 \) and \( \hat{\beta}=0.551 \), which gives  \( U=0.7809 \) and \( p<0.0001 \). For DBN, the estimated \( \hat{\alpha} = 0.011 \), and \( \hat{\beta} = 1.008 \) , which is close to the null values (\( U=-6.26 \times 10^{-5} , p<0.8442 \)).

\begin{figure}
  \includegraphics[width=1\textwidth]{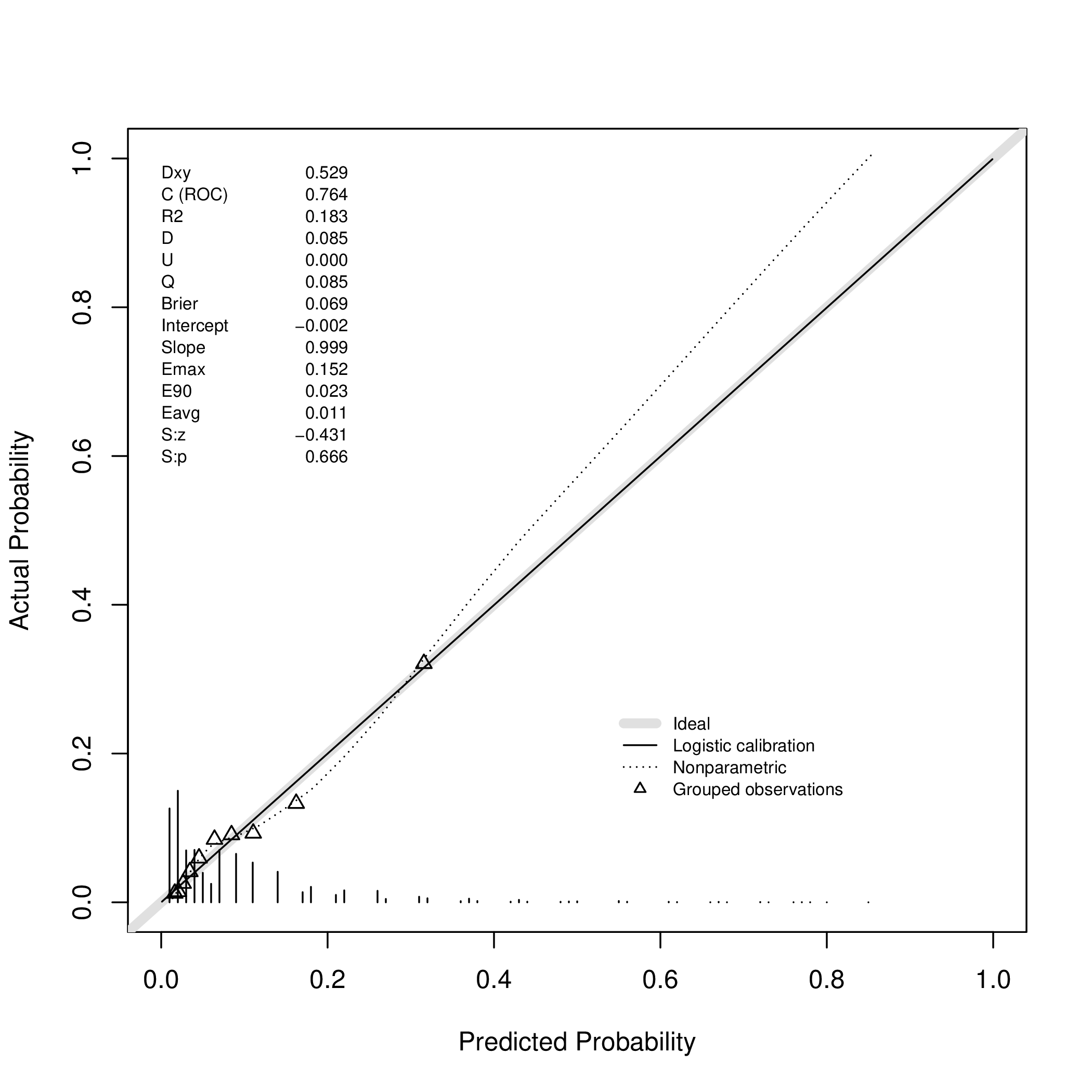}
  \caption{ SOFA first: \( \alpha = -0.003 , \beta = 0.998, U= -8.3 \times 10^{-5}, p=0.9977 \) }
  \label{aFig1}
\end{figure}

\begin{figure}
  \includegraphics[width=1\textwidth]{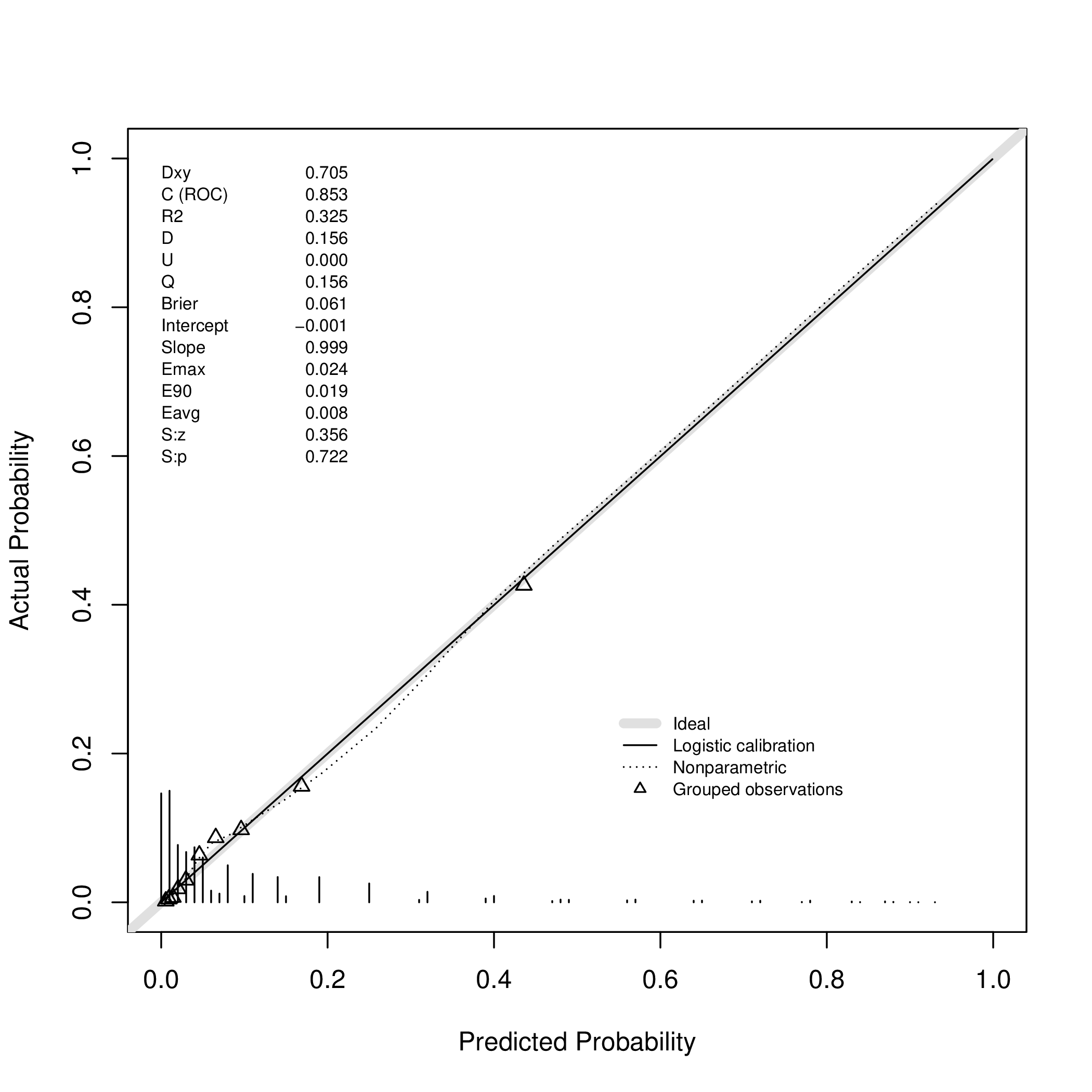}
  \caption{SOFA max: \( \alpha = -0.001 , \beta = 0.999, U= -8.3 \times 10^{-5}, p=0.9993 \) }
  \label{aFig2}
\end{figure}

\begin{figure}
  \includegraphics[width=1\textwidth]{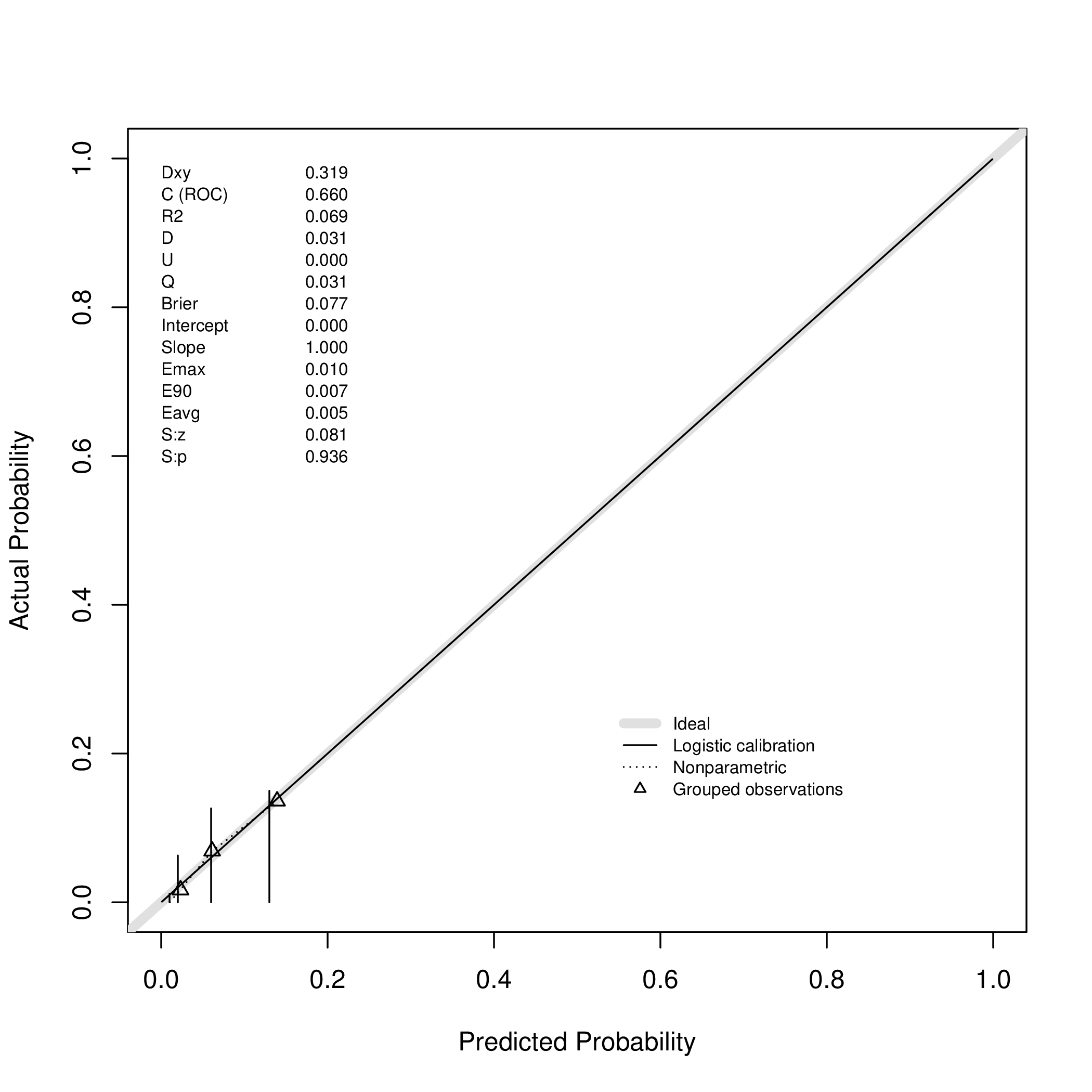}
  \caption{qSOFA: \( \alpha = 0 , \beta = 1, U= -8.2 \times 10^{-5}, p=1.0 \) }
  \label{aFig3}
\end{figure}

\begin{figure}
  \includegraphics[width=1\textwidth]{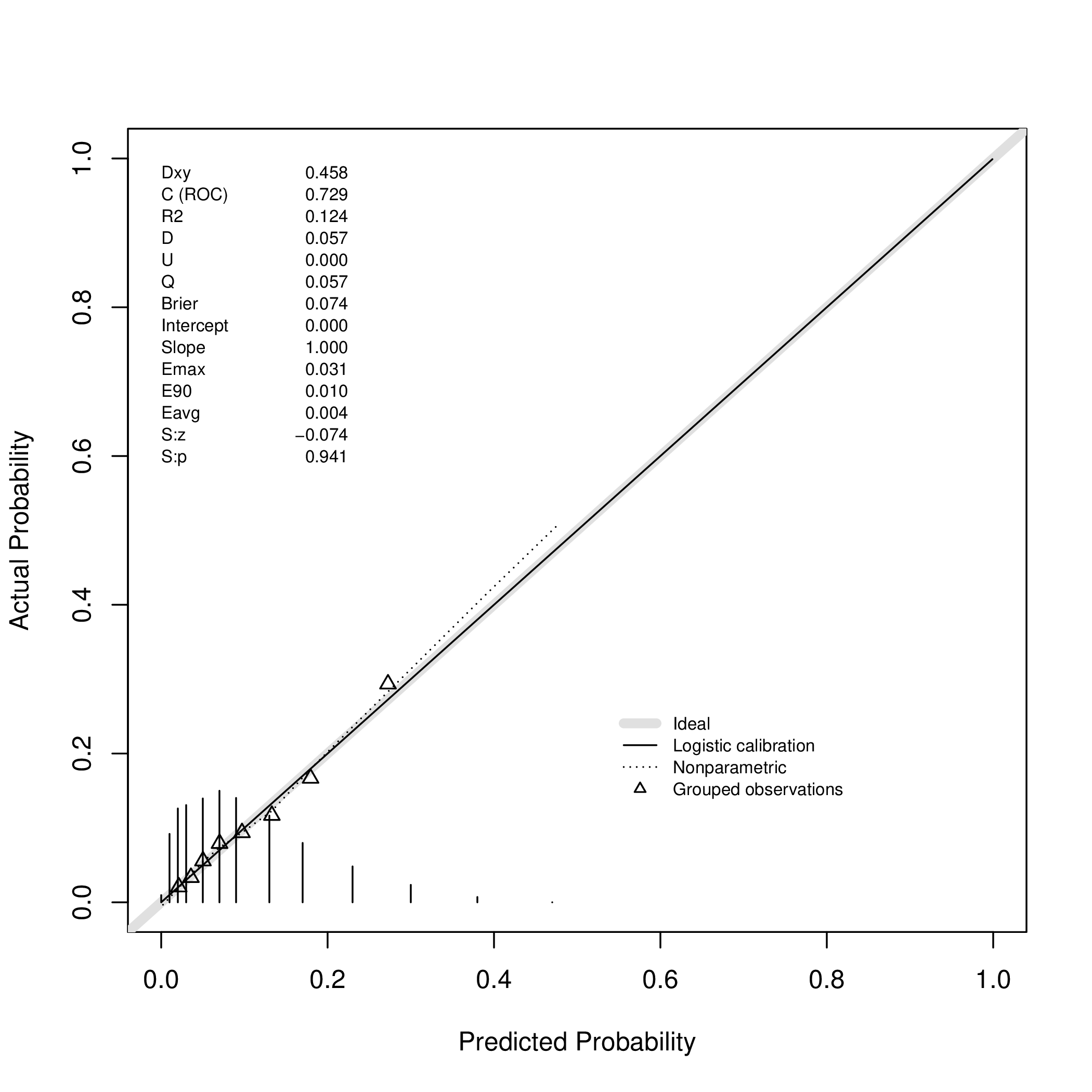}
  \caption{MEWS: \( \alpha = 0 , \beta = 1, U= -8.21 \times 10^{-5}, p=1.0 \) }
  \label{aFig4}
\end{figure}

\begin{figure}
  \includegraphics[width=1\textwidth]{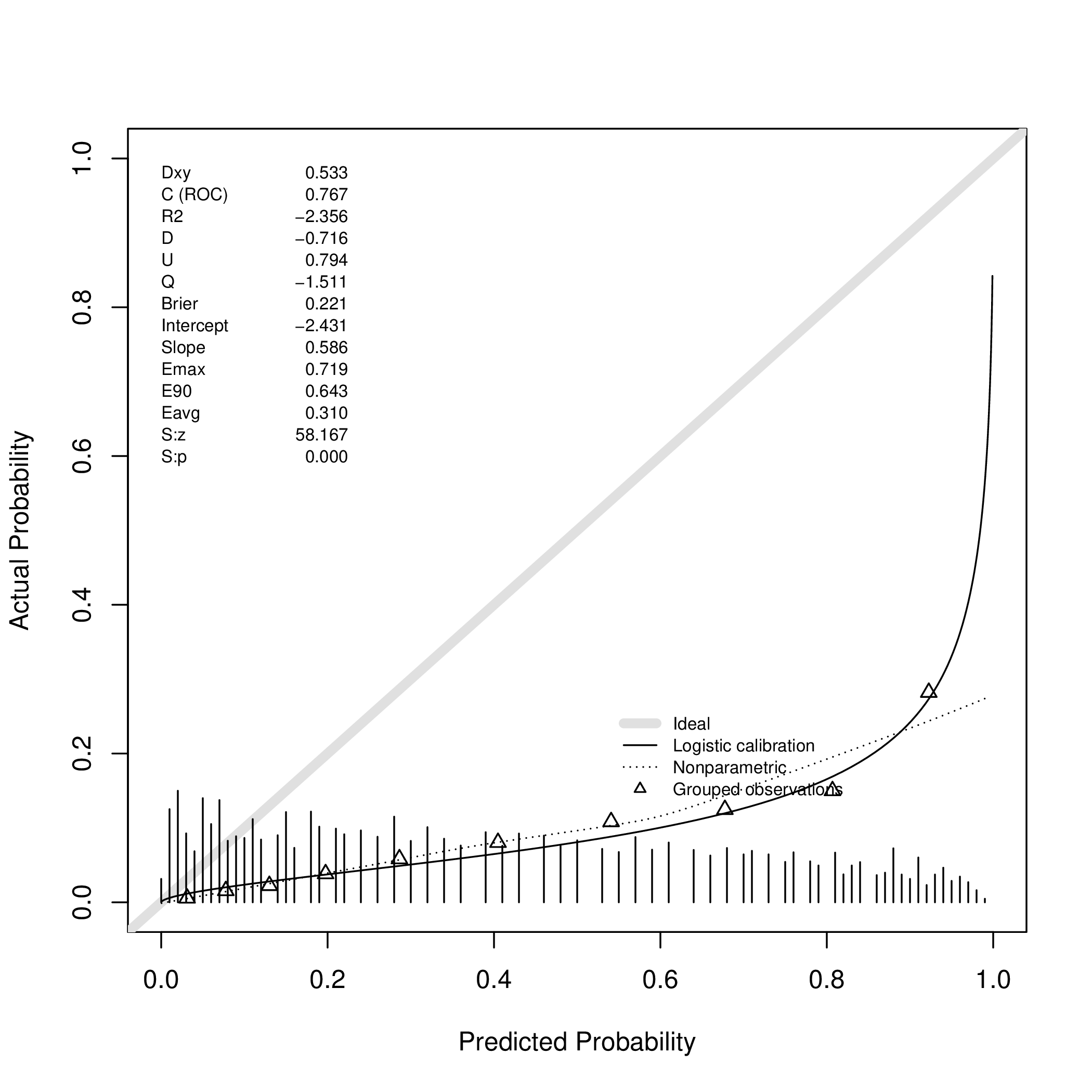}
  \caption{SAPS-II: \( \alpha = -2.431 , \beta = 0.586, U= 0.7943, p=0.0001 \) }
  \label{aFig5}
\end{figure}

\begin{figure}
  \includegraphics[width=1\textwidth]{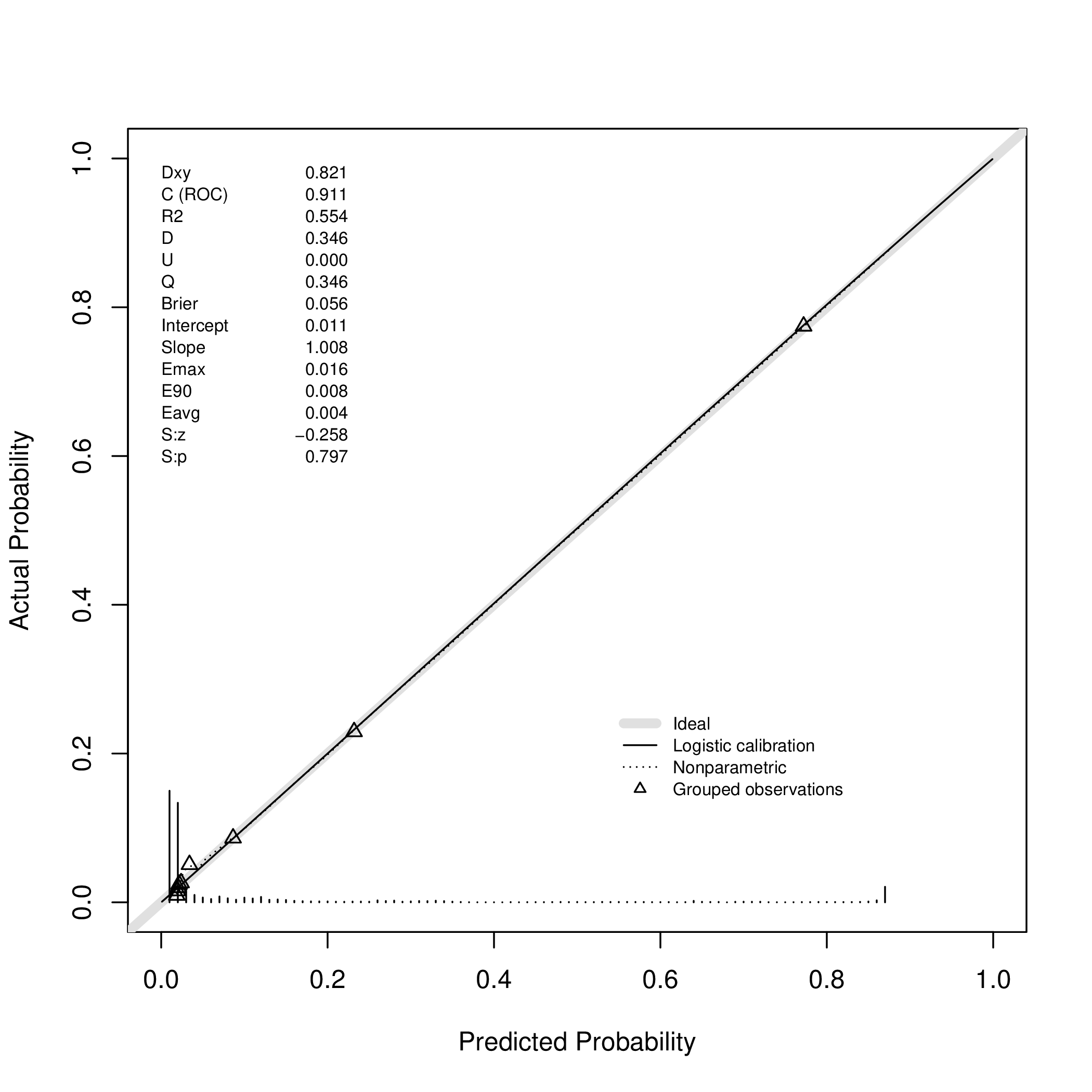}
  \caption{DBN: \( \alpha = 0.011 , \beta = 1.008, U= -7.569 \times 10^{-5}, p=0.8676 \) }
  \label{aFig6}
\end{figure}

\end{document}